\pdfoutput = 1
\documentclass[10pt]{article}
\usepackage{ijcai18}
\usepackage{graphicx}
\usepackage{times}
\usepackage{xcolor}
\usepackage{soul}
\usepackage{amsfonts}
\usepackage{url}
\usepackage{amsmath}
\usepackage{subfigure}
\usepackage[utf8]{inputenc}
\usepackage[small]{caption}

\title{An Encoder-Decoder Framework Translating Natural Language to Database Queries}

\author{
	Ruichu Cai$^1$,
	Boyan Xu$^1$,
	Zhenjie Zhang$^2$,
	Xiaoyan Yang$^2$,
	Zijian Li$^1$,
	Zhihao Liang$^1$
	\\
	$^1$  Faculty of Computer, Guangdong University of Technology, China \\
	$^2$  Singapore R\&D, Yitu Technology Ltd.  \\
	cairuichu@gmail.com,
	hpakyim@gmail.com,
	zhenjie.zhang@yitu-inc.com\\
	xiaoyan.yang@yitu-inc.com,
	leizigin@gmail.com,
	zhihaolzh95@gmail.com 
}
\begin{document}
\renewcommand{\footnotesize}{\normalsize} 
\maketitle

\begin{abstract}
  Machine translation is going through a radical revolution, driven by the explosive development of deep learning techniques using Convolutional Neural Network (CNN) and Recurrent Neural Network (RNN). In this paper, we consider a special case in machine translation problems, targeting to convert natural language into Structured Query Language (SQL) for data retrieval over relational database. Although generic CNN and RNN learn the grammar structure of SQL when trained with sufficient samples, the accuracy and training efficiency of the model could be dramatically improved, when the translation model is deeply integrated with the grammar rules of SQL. We present a new encoder-decoder framework, with a suite of new approaches, including new semantic features fed into the encoder, grammar-aware states injected into the memory of decoder, as well as recursive state management for sub-queries. These techniques help the neural network better focus on understanding semantics of operations in natural language and save the efforts on SQL grammar learning. The empirical evaluation on real world database and queries show that our approach outperform state-of-the-art solution by a significant margin.
\end{abstract}

\section{Introduction}\label{sec:intro}

Machine translation is known as one of the fundamental problems in machine learning, attracting extensive research efforts in the last few decades \cite{koncar1997natural,castano1997machine}. In recent years, with the explosive development of deep learning techniques, the performance of machine translation is dramatically improved, by adopting convolutional neural network \cite{gehring2017convolutional} or recurrent neural network \cite{cho2014learning,wu2016google,zhou2016deep}. 
The growing demands of computer-human interaction in the big data era, however, is now looking for additional support from machine translation to convert human commands into actionable items understandable to database systems \cite{giordani2012translating,li2014constructing,mou2017coupling,popescu2003towards,rabinovich2017abstract,yin2017syntactic}, in order to ease the efforts of human users on learning and writing complicated Structured Query Language (SQL). Our problem is known to be more challenging than the traditional semantic parsing problem, e.g., latest SCONE dataset involving context-dependent parsing \cite{long2016simpler}, because of the high complexity of database querying language. Given a complex real world database, e.g., Microsoft academic database \cite{roy2013microsoft}, it contains dozens of tables and even more primary-foreign key column pairs. A short natural language question, such as ``Find all IJCAI 2018 author names" must be converted into a SQL query with more than 10 lines, because the result query involves four tables.


Recently, a number of research works attempt to apply neural network approaches on data querying, such as \cite{neelakantan2016learning,yin2016neural}, which target to generate data processing results by directly linking records in data tables to the semantic meanings of the natural language questions. There are two major limitations rooted at the design of their solutions. First, such methods are not scalable to big data tables, since the computation complexity is almost linear to the cardinality of the target data tables. Second, the conversion results of such methods are not reusable when a database is updated. The original natural language queries must be recalculated from scratch, in order to generate results on a new table or newly incoming records. The key to a more scalable and extensible solution is to transform original natural language queries into SQL queries instead of query answers, such that the result SQL queries are simply reusable on all tables of arbitrary size at any time.

Technically, we opt to employ encoder-decoder framework as the underlying translation model, based on Recurrent Neural Network and Long Short Term Memory (LSTM) \cite{hochreiter1997long}. Basically, in the encoder phase, the neural network recognizes and maintains the semantic information of the natural language question. In the decoder phase, it outputs a new sequence in another language based on the information maintained in the hidden states of the neural network. Encoder-decoder framework has outperformed conventional approaches over generic translation tasks for various pairs of natural languages. When the output domain is a structured language, such as SQL, although encoder-decoder framework is supposed to learn the grammar structure of SQL when given sufficient training samples, the cost is generally too high to afford. It spends most of the computation power on grammar understanding, but only little effort on the semantical interpretation of original questions. Even given sufficient training data, the output of standard encoder-decoder may not fully comply with SQL standard, potentially ruining the utility of the result SQL queries on real databases.

In this paper, we propose a new approach smoothly combining deep learning techniques and traditional query parsing techniques. Different from recent studies with similar strategy \cite{iyer2017learning,rabinovich2017abstract,yin2017syntactic}, we include a suite of new methods specially designed for structured language outputting. On the encoder phase, instead of directly feeding word representations into the neural network, we inject a few new bits into the memory of the neural network based on language-aware semantical labels over the input words, such as \emph{table names} and \emph{column names} in SQL. These additional dimensions are not directly learnable by language models, but explicitly recognizable based on the properties of the structured language. On the decoder phase, we insert additional hidden states in the memory layer, called grammatical states, which indicate the states of the translation output in terms of Backus-Naur Form (BNF) of SQL. To handle the complexity behind schema-relevant information, our system generates two types of dependency and masking mechanisms to better capture the constraints based on SQL grammar as well as database schema. We also allow the neural network to recursively track the grammatical state when diving into sub-queries, such that necessary information is properly maintained even when nested queries are finished.


The core contributions of the paper are summarized as follows: 1) we present an enhanced encoder-decoder framework deeply integrated with known grammar structure of SQL; 2) we discuss the new techniques included in encoder and decoder phases respectively on grammar-aware neural network processing; 3) we evaluate the usefulness of our new framework on synthetic workload of real world database for natural language querying.



\section{Related Work}\label{sec:related}




The emergence of deep learning techniques, particularly recurrent neural network for sequential domain, enables the machine learning models to build such complicated dependencies, and greatly enhance the translation accuracy. Encoder-decoder framework is known as a typical RNN framework designed for machine translation \cite{cho2014learning,sutskever2014sequence}. On the other hand, convolutional neural network models are recently recognized as an effective alternative to recurrent neural network model for machine translation tasks. In \cite{gehring2017convolutional}, Gehring et al. show that convolution allows the machine learning system to better train translation model by using GPUs and other parallel computation techniques.



In last two years, researchers are turning to adopt recurrent neural network for automatic data querying and programming based on natural language inputs, which aims to translate original natural language into programs and data querying results. Semantic parsing, for example, is the problem of converting natural language into formal and executable logics. In last two years, sequence-to-sequence model is becoming state-of-the-art solution of semantic parsing \cite{xiao2016sequence,dong2016language,guu2017from}. While most of the existing studies exploit the availability of human intelligence for additional labels \cite{jia2016data,liang2016learning}, our approach learns the translation with input-output sample pairs only. While masking is proposed in the literature for symbolic parsing by storing key-variable pairs in the memory \cite{liang2017neural}, the masking technique proposed in this paper supports more complex operations, covering both short-term and long-term dependencies. Moreover, we hereby emphasize that the grammar structure of SQL is known to be much more complicated than the logical forms used in semantic parsing.


Besides of semantic parsing, researchers are also attempting to generate executable logics by directly linking the semantic interpretation of the input natural language and the records in the database. Neural networks are employed to identify appropriate operators \cite{neelakantan2016learning}, while distributed representations are used \cite{mou2017coupling,yin2016neural} to columns, rows and records in the data table. As pointed out in the introduction, such approaches do not scale up in terms of the data size, and the outputs are not reusable over a new data table or updated table with new records.

Recently, \cite{iyer2017learning,rabinovich2017abstract,yin2017syntactic} try to integrate grammar structure into sequence-to-sequence model for data processing query generation. These studies share common idea of our paper on tracking grammar states of the output sequence. Our approach, however, differentiates on two major points. Firstly, we design consistent and systematic approach based on grammar rule (i.e., centered at non-terminal symbols in BNF) for both encoder and decoder phases. Secondly, we include both short-term and long-term dependency in output word screening based on grammar state, exploiting the information from the schemas of the databases. These features bring significant robustness improvement.




\section{Overview}\label{sec:prelim}


\begin{figure}
	\centering
	\includegraphics[width=\columnwidth]{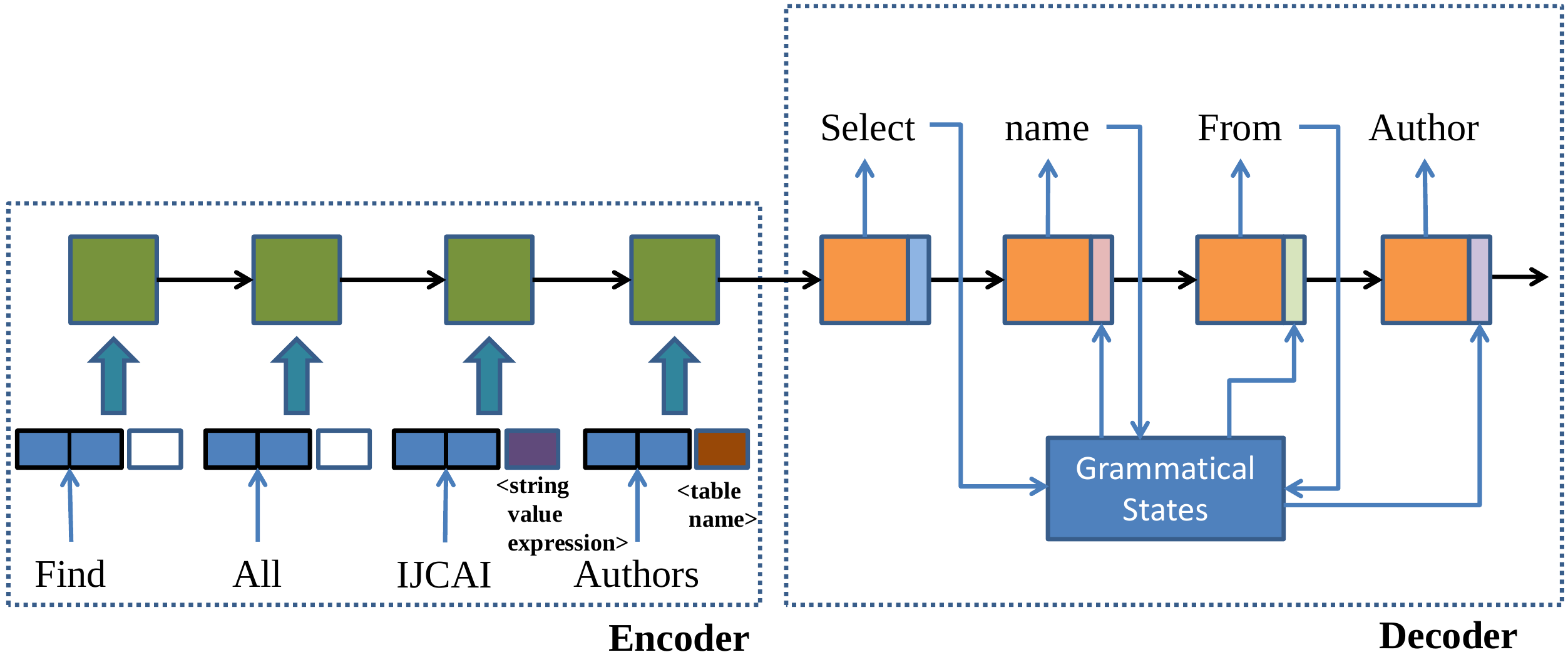}
	\caption{A running example of our new Encoder-Decoder framework. The encoder phase accepts new semantic labels of the input words based on text analysis. The decoder phase employs additional augmented memory controlled by grammatical state.}
	\label{fig:motivation}
\end{figure}

In this paper, we formulate the translation process as a mapping from a natural language domain $\mathbb{N}$ to a structured language domain $\mathbb{S}$, i.e., $\mathbb{N}\mapsto\mathbb{S}$. The input from $\mathbb{N}$ is a natural language sentence, $N=(w_1,w_2,\ldots,w_{L_N})$ with every word $w_i$ from a known dictionary $D_{\mathbb{N}}$. Similarly, the output of the mapping is another text sequence, $S=(w_1,w_2,\ldots,w_{L_S})$, in a structured language domain, e.g., SQL on a relational database, with dictionary $D_{\mathbb{S}}$. The goal of the translation learning is to reconstruct the mapping, based on given samples of the translation, i.e., a training set with natural language and corresponding queries $T=\{(N_1,S_1),\ldots,(N_n,S_n)\}\subset\mathbb{N}\times\mathbb{S}$.


Encoder-decoder framework \cite{sutskever2014sequence} is the state-of-the-art solution to general machine translation problem between arbitrary language pairs. As is shown in Figure \ref{fig:motivation}, there are two phases in the transformation from an input sequence to output sequence, namely \emph{encoder phase} and \emph{decoder phase}. The encoder phase mainly processes the input sequence, extracts key information of the input sequence and appropriately maintains them in the hidden layer, or memory in another word, of the neural network. The decoder phase is responsible for output generation, which sequentially selects output words in its dictionary and updates the memory state accordingly.


In this paper, we propose a variant encoder-decoder model, with new features designed based on the purpose of converting natural language into executable and structured language. The general motivations of these new features are also presented in Figure \ref{fig:motivation}. In the encoder phase, besides of the vectorized representations of the input words, we add a number of additional binary bits into the input vector to the neural network. These binary bits are used to indicate the possible semantical meaning of these words. In our example, the word ``IJCAI'' is labeled as string value expression and the word ``Authors'' is marked as a potential column name in the table. Note that such information is not directly inferrable by a distributed representation system, e.g., \cite{mikolov2013distributed}. In the decoder phase, we also add new binary bits to the hidden memory layer. These states are not manipulated by the neural network, but by certain external control logics. Given the history of the output words, the external logics calculate the grammatical status of the output sequence. These augmented grammatical status is further utilized to mask candidate words for outputting. as well as feedforward features to the hidden layer of neural network. This mechanism enables our system output executable SQL at any time and enhances the learnability of the neural network.


\begin{figure}[t]
	\centering\includegraphics[width=\linewidth]{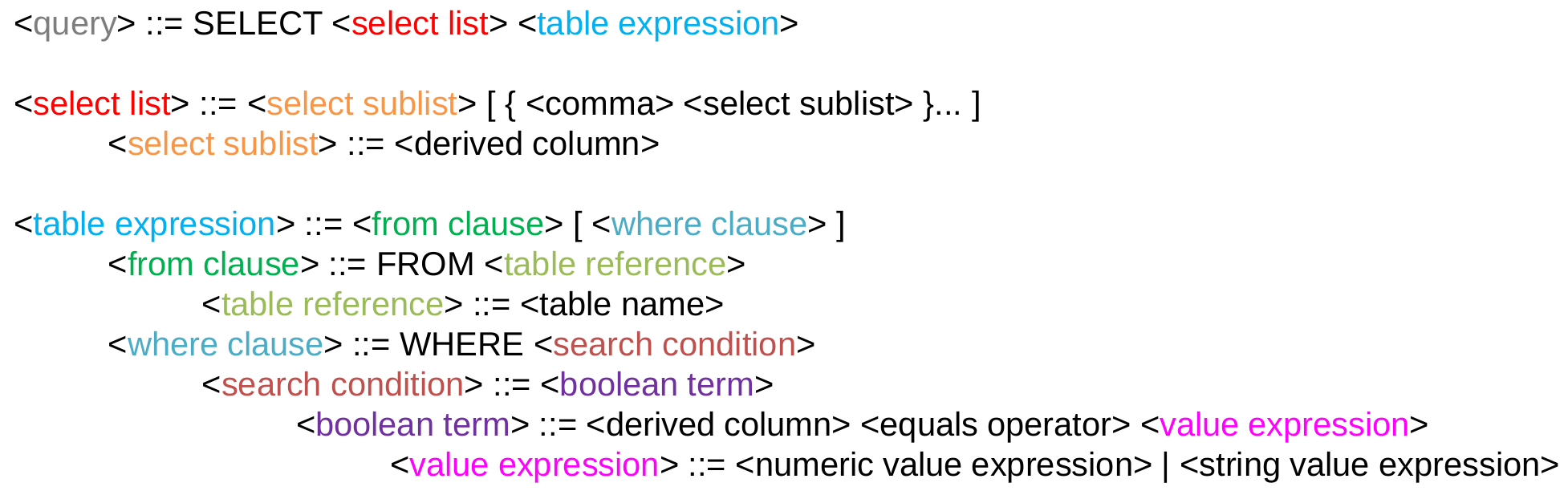}
	\caption{BNF of grammar structure of selection queries in SQL-92 standard. In the derivation rules, we use $\langle\rangle$ to indicate a symbol, $[]$ to indicate an option and $\{\}$ to indicate a block of symbols.}
	\label{fig:bnf}
\end{figure}

Different from recent studies \cite{rabinovich2017abstract,yin2017syntactic}, we utilize Backus Normal Form (BNF) to generate grammatical state tracking. A BNF specification of a language is a set of derivation rules, consisting of a group of \emph{symbols} and \emph{expressions}. There are two types of symbols, \emph{terminal} symbols and \emph{non-terminal symbols}. If a symbols is non-terminal, corresponding expression contains one or more sequences of symbols. These sequences are separated by the vertical bars, each of which is a possible substitution for the symbol on the left. Terminal symbols never appear on the left side of any expression. In Figure \ref{fig:bnf}, we present the BNF of SQL-92, with $\langle$query$\rangle$ as the root symbol. All colored symbols, e.g., $\langle$table expression$\rangle$, are non-terminal symbols, and symbols in black, e.g., $\langle$numeric value expression$\rangle$, are terminal symbols. Theoretically, the language is context-free, if it could be written in form of BNF, and therefore deterministically verified by a push-down automaton. Given the BNF of SQL in Figure \ref{fig:bnf}, parsers in relational database systems can easily track the grammatical correctness of an input SQL query by scanning the query from beginning to end. Although grammar tracking strategy is similar to \cite{rabinovich2017abstract,yin2017syntactic}, we employ short-term and long-term dependencies to accurately mask words based on both SQL grammar and database schema.



\section{Techniques}\label{sec:tech}


\noindent\textbf{Encoder Processing:} The key of encoder phase in the framework is to digest the original natural language input and put the most important information in the memory before proceeding to the decoder phase. In order to extract useful information from the words in the sentence, we propose to extract additional \emph{semantic} features that link the original words to the semantics of the grammatical structure of the target language.

\begin{figure*}[t]
	\centering
	\includegraphics[width=\linewidth]{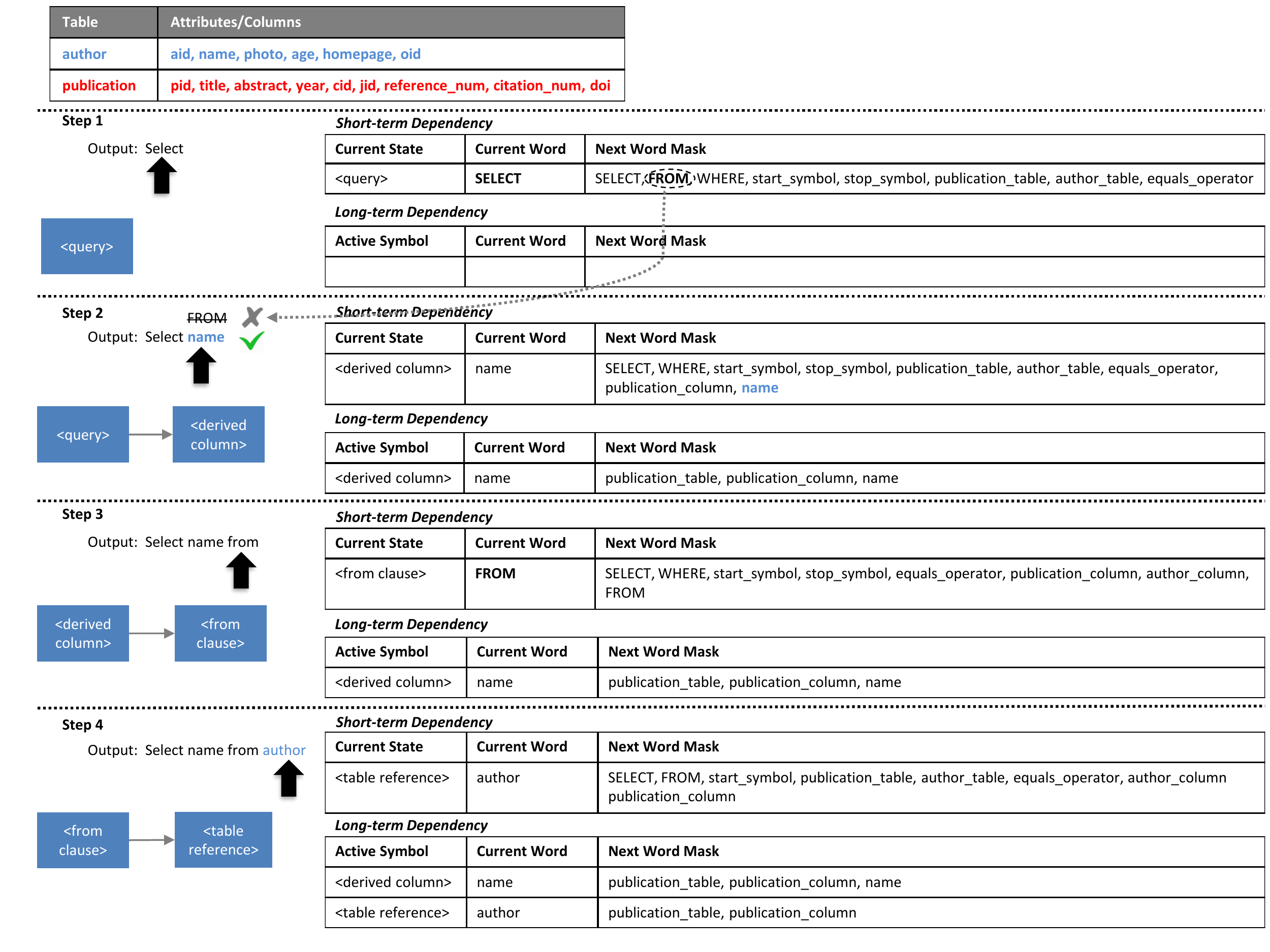}
	\caption{A running example of the decoder phase assuming simple selection query over two tables ``author'' and ``publication'': in the first step, given the grammar state ``Query" and current word ``SELECT", the decoder adds word masking by finding S1, a rule of short-term dependency. The word mask blocks the output of word ``FROM" in next step. After outputting the word ``name", the decoder further adds new word masking by identifying L1 in long-term dependency rules. Because of the word masking from L1, the decoder is only allowed to output ``author" as the table for querying.}
	\label{fig:eg}
\end{figure*}



We generate a group of labels based on the BNF of the target language $\mathbb{S}$. Specifically, each label corresponds to a terminal symbol in the BNF. Based on the BNF in Figure \ref{fig:bnf}, there are four terminal symbols with corresponding labels.

\noindent\textit{$\langle$Derived column$\rangle$}: refers to words used to describe the columns specified in the database query, e.g., the word ``name" in Figure \ref{fig:motivation}.

\noindent\textit{$\langle$Table reference$\rangle$}: refers to words used to describe the tables specified in the database query, e.g., the word "author".

\noindent\textit{$\langle$Value expression$\rangle$}: refers to words containing numeric values used to describe the conditions in the database query.

\noindent\textit{$\langle$String expression$\rangle$}: refers to words containing string values used to describe the conditions in the database query, e.g., the word ``IJCAI" in Figure \ref{fig:motivation}.


Given a small group of samples, we manually label the words with these four label types and employ conditional random fields (CRFs) \cite{lafferty2001conditional} to build effective classifiers for these labels.

\noindent\textbf{Decoder Processing: }We employ two different techniques in the decoder phase, including the embedding of grammar state in the hidden layer and the masking of word outputs. 

Basically, given a particular word in the output sequence, the grammar state of the word is the last expression of BNF this word fits in. When a parser interprets a SQL query, it selects the candidate expression for the words based on the structure of BNF. In the example shown in Figure~\ref{fig:eg}, the parser enters state \emph{derived column} when it encounters word ``name'' in step 2.
To facilitate grammar state tracking, we use a binary vector structure to represent all possible states. The length of the vector is identical to the number of expressions in the BNF of $\mathbb{S}$. Each binary bit in the vector denotes if a particular expression is active based on the parser. When reading a new word of the output of the decoder, the SQL parser updates the grammar vector to reflect the semantic meaning of the word. The grammar state is used not only for state tracking but also for the update of the memory of the neural network. Let $g_t$ denote the grammar status information at time $t$. To incorporate $g_t$ into the model, the memory of the neural network is updated as follows:
\begin{equation}
\begin{aligned}
f_t &= \sigma_g(W_f x_t + U_f h_{t-1} + V_f g_{t-1} + b_f) \\
i_t &= \sigma_g(W_i x_t + U_i h_{t-1} + V_i g_{t-1} + b_i) \\
o_t &= \sigma_g(W_o x_t + U_o h_{t-1} + V_o g_{t-1} + b_o) \\
c_t &= f_t \otimes c_{t-1} + i_t \otimes \sigma_c(W_f x_t + U_f h_{t-1} + V_f g_{t-1} + b_f) \\
h_t &= o_t \otimes \sigma(c_t)\nonumber
\end{aligned}
\end{equation}
where $\otimes$ indicates element-wise multiplication operation.

\begin{table}[t]
	\centering
	
	\small
	\begin{tabular}{|l||l|p{0.9in}||p{0.9in}|}
		\hline
		ID & State & Current Word/Symbol & Next Word/Symbol \\ \hline \hline
		S1 & $\langle$query$\rangle$	& SELECT & $\langle$derived column$\rangle$ \\ \hline
		S2 & $\langle$select list$\rangle$  & $\langle$derived column$\rangle$ & $\langle$comma$\rangle$,FROM \\ \hline
		S3 & $\langle$select list$\rangle$  & $\langle$comma$\rangle$ & $\langle$derived column$\rangle$ \\ \hline
		S4 & $\langle$from clause$\rangle$  & FROM & $\langle$table name$\rangle$ \\ \hline
		S5 & $\langle$from clause$\rangle$  & $\langle$table name$\rangle$ & WHERE, Stop\_symbol \\ \hline
		S6 & $\langle$where clause$\rangle$ & WHERE & $\langle$derived column$\rangle$ \\ \hline
		S7 & $\langle$where clause$\rangle$ & $\langle$derived column$\rangle$   & $\langle$equals operator$\rangle$ \\ \hline
		S8 & $\langle$where clause$\rangle$ & $\langle$equals operator$\rangle$  & $\langle$value expression$\rangle$ \\ \hline
		S9 & $\langle$where clause$\rangle$ & $\langle$value expression$\rangle$ & Stop\_symbol \\ \hline
	\end{tabular}
	
	\caption{Partial rules of \emph{Short-term Dependencies}.}\label{tb:short}
\end{table}

\begin{table}[t]
	\centering
	
	\small
	\begin{tabular}{|l||p{0.9in}|p{0.4in}||p{1.25in}|}
		\hline
		ID & Symbol & Current Word&  Long Term Word Mask \\ \hline
		L1 & $\langle$derived column$\rangle$ & name & publication\_table/column, name
		\\ \hline
		L2 & $\langle$table name$\rangle$ & author & publication\_table/column
		\\    \hline
	\end{tabular}
	
	\caption{Partial rules of \emph{Long-term Dependencies}.}\label{tb:long}
\end{table}

\noindent\textbf{Output Words Masking: }In the decoder, there are two types of word masks used to filter out invalid words for outputing, which are mainly based on \emph{short-term dependencies} and \emph{long-term dependencies} respectively. At each step, the decoder chooses one rule from candidate short-term dependencies, e.g., rules in Table \ref{tb:short}, and possibly multiple rules from candidate long-term dependencies, e.g., rules in Table \ref{tb:long}. The short-term dependency rule is updated according to the current grammar state as well as the last output word from the decoder. In Table \ref{tb:short}, the columns of ``State" and ``Current Word/Symbol" are used for rule matching, while the column ``Next Word/Symbol" indicates all valid output words in next step of the decoder. Once the decoder identifies a matching rule, it generates a mask on the dictionary to block the output of words not allowed by the rule. Long-term dependencies are updated based on the active symbols chosen by the SQL parser, maintained in the grammar state vector. For each active symbol, the decoder includes a rule from all long-term dependency rules, e.g., Table \ref{tb:long}, by matching on ``Symbol" and "Word". Given the rule, the decoder generates the output word mask accordingly. The rules for long-term dependencies are removed from the decoder, only when the corresponding symbol turns inactive.


We use a binary vector $s$ to indicate the masks generated by the single rule of short-term dependency, and $l_{i}$ for the $i$-th mask generated by the rule of long-term dependencies. Given these masks, the word selection process in the decoder is modified as:
\begin{eqnarray}
y_t = \sigma_y(W_y h_t + b_y) \otimes s \otimes l_1 ... \otimes l_L,
\end{eqnarray}
%
where $L$ is the number of active long-term dependency rules.

In Figure \ref{fig:eg}, we present a detailed running example on the evolution of the active rules and corresponding masks, to elaborate the effect of combining the neural network and the grammar state transition. Following the example in Figure \ref{fig:motivation}, the query attempts to retrieve names from the author table, with the grammar states and masks updated based on the descriptions above.

\noindent\textbf{Dependency Rule Generation: }The automatic generation of rules for short-term dependencies and long-term dependencies are different. Due to the limited space, we only provide a sketch of the generation methods in the current version.

For short-term dependency, the framework identifies the reachable terminal symbols for every pair of symbol and word. Consider S1 in Table \ref{tb:short}. Given the symbol $\langle$query$\rangle$ and word output ``SELECT", the only matching expression in BNF is $\langle$query$\rangle$ ::= SELECT $\langle$select list$\rangle$ $\langle$table expression$\rangle$. The following symbol is $\langle$select list$\rangle$. Since $\langle$select list$\rangle$ is not a terminal symbol, we iterate over the BNF to find the terminal symbols to generate in next step. In this case, we reach the terminal symbol $\langle$derived column$\rangle$ and thus insert it into the fourth column of S1 in Table \ref{tb:short}.

For long-term dependency, the framework must combine the BNF as well as the schema of the database. Currently, we only consider $\langle$derived column$\rangle$ and $\langle$table name$\rangle$, which forbid the adoption of non-relevant tables and columns in the rest of the SQL query.




\section{Experiments}\label{sec:exp}

\noindent \textbf{Workload Preparation:} We run our experiments on three databases, namely \emph{Geo880}, \emph{Academic} and \emph{IMDB}. The workload on \emph{Geo880} is generated by converging logical form outcomes to equivalent relational table and SQL queries. \emph{Academic} database has 17 tables,
collected by Microsoft Academic Search \cite{roy2013microsoft}. This database is employed in the experiments of ~\cite{li2014constructing}. \emph{IMDB} has 3 tables, containing records of 3,654 movies, 4,370 actors and 1,659 directors. On \emph{Academic} and \emph{IMDB}, we generate SQL query workloads and ask volunteers to label the queries with natural language descriptions. Specifically, two types of workloads are generated, namely \emph{Select} workload and \emph{Join} workload. The queries in \emph{Select} and \emph{Join} (with 4 concrete aggregation operators for AGG, including Min, Max, Average and Count) are in the following two forms respectively:

\begin{center}
	\begin{verbatim}
	SELECT <column_array>
	FROM <table> WHERE <column> =/> <value>
	\end{verbatim}
\end{center}


\begin{center}
	\begin{verbatim}
	SELECT AGG(<table_1.column_array>)
	FROM <table_1> INNER JOIN <table_2>
	ON <table_1.key> = <table_2.key>
	WHERE <table_2.column> =/> <value>
	\end{verbatim}
\end{center}


Given the standard forms of the queries above, we generate concrete queries by randomly selecting the tables and columns without replacement. For each combination of tables and columns, we randomly select values for the conditions in the queries. By manually filtering out meaningless queries, we generate 35 queries on \emph{Academic} database and 75 queries on \emph{IMDB} database. Each query is manually labeled by at least 5 independent volunteers. Given a pair of natural language description label and query, we further generalize it to a group of variant queries, by modifying the search conditions in where clauses. Consequently, we get 1,456 (376 select query and 1,080 join query) pairs of samples on \emph{Academic} database and 2,103 (1,082 select query and 1,021 join query) pairs of samples on \emph{IMDB} database. We also build a \emph{Mixed} workload, by simply combining all samples from both \emph{Select} and \emph{Join} workloads. We reuse the queries and natural language descriptions in \emph{Geo880} database, and use the standard training/test split as in~\cite{iyer2017learning}.



\noindent \textbf{Baseline Approaches:} We employ two state-of-the-art and representative approaches as baseline in our experiments, including NLP translation model \emph{NMT} \cite{wu2016google} and semantic parsing model with feedback \emph{SPF} \cite{iyer2017learning}.
Note that we do not compare against cell-based data querying approaches \cite{neelakantan2016learning,yin2016neural}, because they are only applicable to small tables while our testing databases contain way tens of thousands records.

\noindent \textbf{Performance Metrics:} We examine the quality of translation using three types of metrics. First, we report the token-level BLEU following~\cite{yin2017syntactic} to measure the quality of translation. Second, given the groundtruth SQL query $q$ and the predicted one $\widehat{q}$, we measure \emph{query accuracy} as the fraction of queries with identical returned tuples. This is assessed by executing the predicted and groundtruth queries in the databases and examine their returned tuples.
Third, we calculate the \emph{tuple recall} and \emph{tuple precision} of returned tuples of each $\widehat{q}$, by comparing these tuples against the outcomes from groundtruth $q$. The average precision and recall over all queries are reported. Note that the second metric focuses on query-level correctness, while the third metric evaluates individual tuples in query results. They are therefore numerically independent. All numbers reported in the experiments on \emph{MAS} and \emph{IMDB} are average of 5-fold cross validations.

%

\noindent \textbf{Model Training and Optimization:} In preprocessing, our approach uses NLTK to implement Conditional Random Fields~\cite{lafferty2001conditional} (CRFs) to annotate the natural language queries. The overall accuracy of annotation result is over 99.5\%. Therefore, the semantic features of the input words fed to the encoders are highly reliable. Our model is implemented in Tensorflow 1.2.0. The distributed representations of the words in the dictionary are automatically calculated and optimized by Tensorflow. We optimize the hyerparameters in all approaches and use the configuration with best results. The result hyperparameters are listed in Table~\ref{tab:exp:parameter}.

\begin{table}
	\centering
	\small
	
	\begin{tabular}{|l|c|c|c|}
		\hline
		Hyperparameter    & NMT & SPF & Ours \\ \hline \hline
		Batch Size        & 128 & 100 & 128 \\ \hline
		Hidden Layer Size & 512 & 600 & 512 \\ \hline
		Encoder Layer     &	2   & 2   &	2 \\ \hline
		Decoder Layer	  & 2	& 1	  & 2 \\ \hline
		Optimizer &	ADAM & ADAM & ADAM \\ \hline
		Learning Rate &	0.001 &	0.001 & 0.001 \\ \hline
		Bidirectional Encoder &	Used &	Used & Used \\ \hline
		Encoder Dropout Rate &	0.2 &	0.4 & 0.2 \\ \hline
		Decoder Dropout Rate &	0.2 &	0.5	& 0.2 \\ \hline
		Beam Search Size	& -	& 5	& - \\
		\hline
	\end{tabular}
	
	\caption{Hyerparameters of all approaches in experiments.}\label{tab:exp:parameter}
\end{table}

\begin{table}[!t]
	\centering
	\small
	
	\begin{tabular}{|l|c|c|c|}
		\hline
		Metric          & NMT  & SPF  & Ours \\ \hline \hline
		BLEU            & 83.2 & 38.1 & \textbf{85.2} \\ \hline
		Query Accuracy  & 75.0 & 81.7 & \textbf{82.8} \\ \hline
		Tuple Recall    & 77.4 & 83.7 & \textbf{84.1} \\ \hline
		Tuple Precision & 76.9 & 83.6 & \textbf{83.7} \\ \hline
	\end{tabular}
	
	\caption{Results on \emph{Geo880} workload.}\label{tab:exp:geo}
\end{table}

\begin{table}[!t]
	\centering
	\small
	
	\begin{tabular}{|l|c|c|c|}
		\hline
		Metric          & NMT  & SPF  & Ours \\ \hline \hline
		BLEU            & 82.6 & 82.8 & \textbf{83.0} \\ \hline
		Query Accuracy  & 43.8 & 45.5 & \textbf{47.9} \\ \hline
		Tuple Recall    & 62.7 & 64.6 & \textbf{66.2} \\ \hline
		Tuple Precision & 63.8 & 65.1 & \textbf{66.6} \\ \hline
	\end{tabular}
	\caption{Results on \emph{MAS} workload.}\label{tab:exp:mas}
\end{table}

\begin{table}
	\centering
	\small

	\begin{tabular}{|l|c|c|c|}
		\hline
		Metric          & NMT  & SPF  & Ours \\ \hline \hline
		BLEU            & 85.7 & \textbf{86.7} & 85.7 \\ \hline
		Query Accuracy  & 91.7 & 95.4 & \textbf{97.2} \\ \hline
		Tuple Recall    & 96.9 & \textbf{97.8} & 97.5 \\ \hline
		Tuple Precision & 96.9 & \textbf{97.5} & \textbf{97.5} \\ \hline
	\end{tabular} 
	\caption{Results on \emph{IMDB} workload.}\label{tab:exp:imdb}   
\end{table}

\begin{table}
	\centering
	\small
	
	\begin{tabular}{|l|c|c|c|}
		\hline
		Metric & - Short & - Long & - State \\ \hline \hline
		BLEU & 0.37 & -0.21 & 0.60 \\
		Query Accuracy & -1.82 & -1.82 & -0.91 \\
		Tuple Recall & -2.15 & -1.57 & -1.58 \\
		Tuple Precision & -1.97 & -2.10 & -1.52 \\
		\hline
	\end{tabular}
	
	\caption{Evaluation of individual component on training set of \emph{Geo880} workload using 5-fold cross-validation. Difference between the ``simpler'' model and our original one are reported.}\label{tab:exp:component}
\end{table}

\noindent\textbf{Experimental Results:} We report the experimental results on three databases in Tables \ref{tab:exp:geo}, \ref{tab:exp:mas} and~\ref{tab:exp:imdb} resepctively. In terms of translation quality, our model achieves the highest BLEU on \emph{Geo880} and \emph{MAS} while SPF performs the best on IMDB. The BLEU of SPF on Geo880 (38.1) is much lower than that of the other methods. This is because SPF uses templates to enlarge the training data significantly. Thus it tends to generate queries following those templates, which although returns the identical results but contains redundant components in the predicted query.
In terms of quality of returned tuples by predicted queries, our model achieves the highest query accuracy on all three databases, i.e., the highest percentage of predicted queries with identical returned tuples. It is a significant improvement over the existing methods. The system could return completely right answers to over 80\% of the questions on \emph{Geo880} and \emph{IMDB} databases. Although the query accuracy of all approaches is below 50\% on \emph{MAS} database, due to the high complexity on both schema and content, the recall and precision of the outcomes are all above 60\%. It implies that there remains certain utility even when the translation results contain errors.
An interesting observation on the results over \emph{IMDB} database is: although SPF achieves the highest BLEU, the accuracy on query results by SPF and our approach are almost identical. It shows that translation quality, as used as the golden standard in traditional machine translation tasks, may not be the best metric for our problem setting.

We conduct ablation studies on the training set of \emph{Geo880} (Table~\ref{tab:exp:component}) and find that short/long-term dependencies (Short/Long) and grammar state (State) help improve quality of returned tuples in terms of query accuracy and tuple recall/precision. However,  short-term dependencies and grammar state have negative effect on BLEU, i.e., the predicted queries are more similar to the groundtruth in the token level but are less accurate. This further implies that BLEU may not be the best metric for our problem setting.

\section{Conclusion}\label{sec:conclu}
In this paper, we present a new encoder-decode framework designed for translation from natural language to structured query language (SQL). The core idea is to deeply integrate the known grammar structure of SQL into the neural network structure used by the encoders and decoders. Our results show significant improvements over baseline approaches for standard machine translation, especially on the accuracy of outcomes by executing the SQL queries on real databases. It greatly improves the usefulness of natural language interface to relational databases.

Although our technique is designed for SQL outputs, the proposed techniques are generically applicable to other languages with BNF grammar structures. As future work, we will extend the usage to automatic programming, enabling machine learning systems to write programs, e.g., in C language, based on natural language inputs.

\section*{Acknowledgements}\label{sec:acknow}
This research was supported in part by NSFC-Guangdong Joint Found (U1501254), Natural Science Foundation of China (61472089), Guangdong High-level Personnel of Special Support Program (2015TQ01X140), Pearl River S\&T Nova Program of Guangzhou (201610010101), Science and Technology Planning Project of Guangzhou (201604016075), Natural Science Foundation of Guangdong (2014A030306004, 2014A030308008), Science and Technology Planning Project of Guangdong (2015B010108006, 2015B010131015). 
And it was done when Zhang and Yang were with Advanced Digital Sciences Center, supported by the National Research Foundation, Prime Minister’s Office, Singapore under its Campus for Research Excellence and Technological Enterprise (CREATE) programme.


\bibliographystyle{ijcai18}
\bibliography{ijcai18-translation}

\end{document}